# Shifting Trends of COVID-19 Tweet Sentiment with Respect to Voting Preferences in the 2020 Election Year of the United States

Megan Doman      Jacob Motley      Hong Qin      Mengjun Xie      Li Yang
Department of Computer Science and Engineering, College of Engineering and Computer Science
University of Tennessee at Chattanooga, Chattanooga, Tennessee, U.S.A.
syn472@mocs.utc.edu      yqs872@mocs.utc.edu      hong-qin@utc.edu      mengjun-xie@utc.edu      li-yang@utc.edu

*Abstract*—**COVID-19 related policies were extensively politicized during the 2020 election year of the United States, resulting in polarizing viewpoints. Twitter users were particularly engaged during the 2020 election year. Here we investigated whether COVID-19 related tweets were associated with the overall election results at the state level during the period leading up to the election day. We observed weak correlations between the average sentiment of COVID-19 related tweets and popular votes in two-week intervals, and the trends gradually become opposite. We then compared the average sentiments of COVID-19 related tweets between states called in favor of Republican (red states) or Democratic parties (blue states). We found that at the beginning of lockdowns sentiments in the blue states were much more positive than those in the red states. However, sentiments in the red states gradually become more positive during the summer of 2020 and persisted until the election day**.

## I. INTRODUCTION

The year 2020 saw a wide variety of government-issued responses to the outbreak of the SARS-CoV-2 pandemic in the United States. It has been acknowledged by researchers that these measures were quickly politicized and highly partisan. For example, Gusmano et al. reported the disparity between measures taken by Democrat state leaders, suggesting that these tended to be more immediate based on suggestions by federal health organizations, as opposed to Republican state leaders, who it is said were likely to take less restrictive measures based on the cues of the Republican sitting president at the time [1]. Though it is evident that politics played a significant role in implementation and adherence to measures regarding public health and safety, there are other factors to consider that would play into how these issues were discussed throughout the year leading up to the presidential elections in November. For example, Harvey discussed the effect of lockdown fatigue, or the role of stress involved in the isolation and uprooting of daily life because of COVID-19 lockdown measures, which led to discontent and reduced adherence to isolation measures as time went on [2]. This would suggest the expectation of changing trends in sentiment as people became more uncomfortable with the lockdown measures. Particularly in a discussion of Twitter discourse during the pandemic, it has been observed that social media overuse during lockdown may have contributed to this fatigue, such as the negative effect that the constant stream of information had on the generation Z age cohort [3]. This means that the data gathered from social media can be particularly valuable as it was not just a reflection of users' views but also that it impacted the users who read it during this period, thereby exerting influence on the wider population. This is important as it ties into the spread of misinformation, which has also been a widely documented issue of politics, especially during the pandemic. Social media such as Twitter has been a hotspot for the quick spread of misinformation, such that the website set up resources to fact-check many posts related to COVID-19 [4]. This means that Twitter has been accepted as a source of political information by many of its users, making more evident the value of social media analysis as a line of inquiry regarding politics and the pandemic.

For the United States, the red and blue states refer to states whose voters predominantly choose the candidates from the Republican Party or Democratic Party, respectively [5]. Here, we wanted to investigate if it is possible that analyzing the sentiment had in online discussion of the politicized issues in different areas of the country would give insight into the local political leanings when it came time to vote in the 2020 presidential election. Our goal is that the information gathered here can have applications in politics, allowing for the ongoing analysis of the online discourse of issues as a more efficient and less selective method of gauging constituent interest rather than polling individuals and that these results can potentially be used to understand and hopefully reevaluate the role of politics as it is leveraged even in the face of a major emergency such as a global pandemic.

Twitter sentiment analysis has previously been used to track public opinion over an election cycle, such as the tracking of responses to individual events in the 2012 presidential races [6]. Our research focuses particularly on the effect of pandemics on a time of emergency, which is significant source of intrigue in the social sciences for how they expedite change and magnify issues. This means that the political issues of the election would have been much more pressing than in a year of less hardship for the general populace. Our comparison of these polarizing issues will contribute to the existing literature on the study of sociological and political impacts of the COVID-19 pandemic.

## II. MATERIALS AND METHODS

This study used publicly available popular voting data broken down by state and party and published location-tagged daily Twitter sentiment data.

### A. Twitter Data

Analysis was performed on a dataset of Twitter posts collected based on a selection of key terms related to the COVID-19 pandemic [7]. Collection within this dataset started on March 20, 2020, just at the beginning of states issuing stay-at-home orders and lockdown procedures [8]. We ended our analysis on the date of the general elections, November 4, 2020.

### B. Voting Data of the 2020 U.S. Election

Popular voting data arranged by state was obtained from a public online nonpartisan source [9]. This data set was imported and indexed by state abbreviation. For this study, we parsed out only the quantitative popular voting data and which party was called for each state.

### C. Sentiment Analysis

Sentiment analysis was performed using VADER Sentiment Analysis, a tool designed with social media posts in mind in order to better attune the expected input [10]. VADER utilizes natural language processing (NLP), machine learning methods and five generalizable heuristic rules to assign each text a sentiment score between -1 and 1, representing perfectly negative and perfectly positive sentiment intensity, respectively. Retweets and duplicated tweets were excluded from sentiment estimation. We estimated the daily average sentiment intensity for each state in the United States, which then was imported to Jupyter Notebook using the pandas package. The dates were parsed as the indices to generate the data frames.

### D. Data Processing

We removed any locations that were not shared between both datasets and transposed the sentiment dataset such that the indices are the state abbreviations, allowing for the concatenation of these data frames by row. Of the voting data, we removed every column except for the state abbreviation indices, the qualitative data of which party won that state, and the percentage of the vote for Democrats, Republicans, and other. We combined the data frames by index, the state abbreviation, and turned the percentages into decimals such that the voting data is now on a similar scale to the sentiment data (Figure 1).

|    | called | dem_percent | rep_percent | other_percent | 2020-03-20 | 2020-03-21 | 2020-03-22 |
|----|--------|-------------|-------------|---------------|------------|------------|------------|
| TX | R      | 0.465       | 0.521       | 0.015         | 0.156886   | 0.158173   | 0.279156   |
| CA | D      | 0.635       | 0.343       | 0.022         | 0.159037   | 0.322140   | 0.253292   |
| CO | D      | 0.554       | 0.419       | 0.027         | 0.632033   | 0.467900   | 0.291575   |
| IL | D      | 0.575       | 0.406       | 0.019         | 0.220340   | 0.194075   | 0.423533   |
| HI | D      | 0.637       | 0.343       | 0.020         | 0.000000   | 0.000000   | 0.000000   |

Figure 1 - Sample of combined data frame. Average COVID-19 tweet sentiment for each state is estimated daily.

### E. Correlation and Heatmap

Python Package Seaborn [11] was used to calculate the matrix of correlation coefficients and generate the heatmaps comparing correlation between national popular vote percentages by party and average Twitter sentiment during the corresponding time periods. We trimmed this data frame to dates along the y-axis and percentages along the x-axis for ease of viewing and comprehension.

### F. Sliding-time-window analysis

Daily averages of tweet sentiment are highly noisy. We therefore applied a 14-day sliding window to estimate the two-week trend. In order to compare the tweet sentiment intensity between blue and red states, we estimate a ratio between the average sentiment for Democrat-called versus Republican-called states in the sliding time windows.

## III. RESULTS

### A. Shifting correlations of COVID-19 tweet sentiment with voting preferences

In order to examine potential association of COVID-19 tweets with voting preferences, we performed a two-week sliding window analysis. In each two-week window, we estimate the correlation coefficient between the average tweet sentiment intensity and the percentage of votes for Democratic, Republican, and other parties.

We visualize these two-week sliding window correlation results in a heatmap (Figure 2). A Positive correlation is represented by intensity of red color, whereas negative correlation is represented by intensity of blue color. The center of the color representation is near gray color, corresponding to a coefficient of zero.

Because votes for the other parties at 1.8% are an extremely small fraction, we expect that Democratic and Republican votes would correlates with average tweet sentiment in oppositive ways. For example, in each row of Figure 2, a red cell in the column of Democratic voting percentage often correspond to a blue cell in the column of Republican voting percentage.

When examining the correlation results from March to November (Figure 2), we can observe a shift that is occurring from mid-April to late May of 2020. Before mid-April, there are generally weak positive correlations between average tweet sentiment with Democratic voting percentages. After May, there are generally weak negative correlations between average tweet sentiment with Democratic voting percentages. This shift can be verified in the column for Republican voting percentage, expect that the color pattern changes in the opposite direction (Figure 2).

Overall, we can observe that positive correlations between tweet sentiments and Republican voting percentages occurred in more two-week periods before the election. It is noteworthy that the incumbent presidential candidate was from the Republican party during the 2020 election.

were generally more positive in the blue states than in the red states. The ratio of blue versus red state sentiment intensities has a declining pattern during the summer of 2020, and generally stays below the gray horizontal line of 1 from the summer to the election day.

Hence, the changing pattern of the blue-versus-red ratio in Figure 3 is consistent with the correlation heatmap in Figure 2. Both figures shows that COVID-19 tweet sentiments were initially more positively associated with Democratic voting preference at the beginning of the pandemic, but this correlation declined after May 2020. Gradually, the positive intensity of COVID-19 tweets associated with Republican voting preference slightly overtook that with Democratic voting preference.

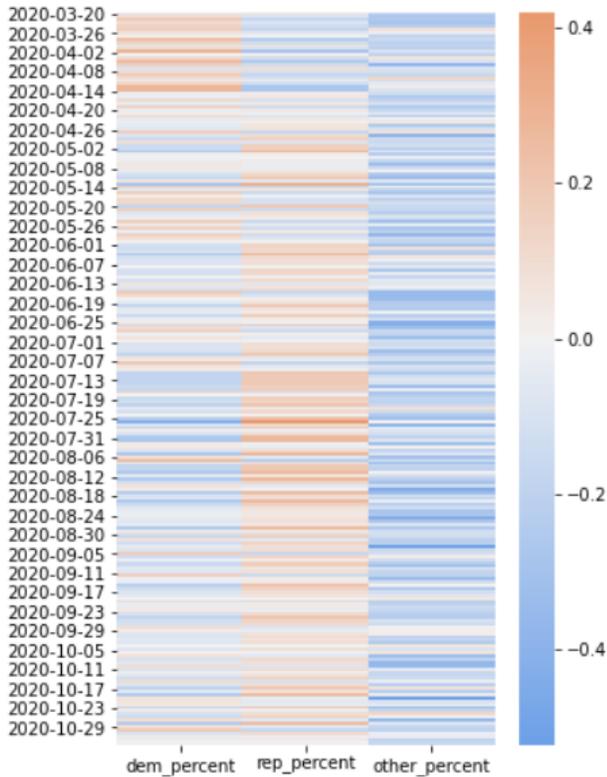

Figure 2 - Shifting correlations of average sentiment intensities of COVID-19 tweets with the percentages of votes for Democratic (dem_percent) and Republican (rep_percent). Vote for other party is represented by other_percent. Correlation analyses were performed in two-week sliding windows.

*B. Switching of relative sentiment of COVID-19 tweets in blue and red states over time*

To further investigate possible association of COVID-19 tweets with the 2020 voting results, we compared the trend of COVID-19 tweet sentiment in blue and red states over time during the 2020 election year (Figure 3). The blue states are the states where the Democratic presidential candidate was declared winner based on a simple majority, and the red states are those where the Republican presidential candidate was declared winner. No state was won by other parties other than the Democratic or Republican parties.

In order to discern the short-term trends, we chose to use a sliding window technique to mitigate the daily fluctuating noises of tweet sentiment. We estimated the average sentiments of COVID-19 tweets in a two-week sliding window from March to November 2020, in blue and red states, respectively.

In order to highlight the relative change over time between the blue and red states, we estimate the ratio of average tweet sentiment intensity in the blue states versus that in the red states in each sliding window. To illustrate the difference between blue and red states, we add a gray horizontal line corresponding to a ratio of 1 in Figure 3. It can be observed that, from March to May, the average sentiment intensities of COVID-19 tweets

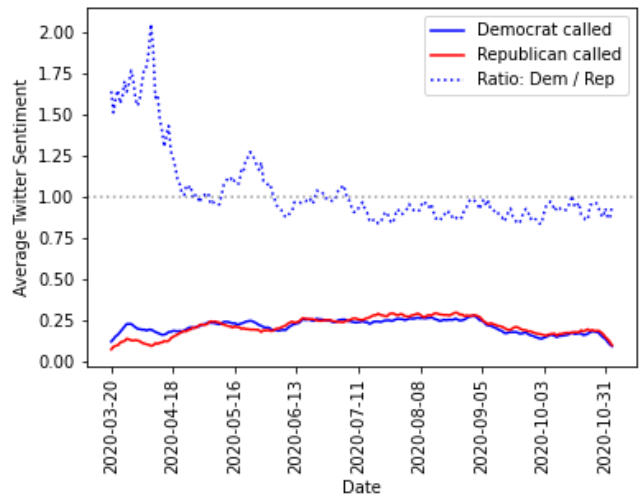

Figure 3 – Shifting trend of relative sentiment intensities of COVID-19 tweets in blue and red states over time. Average sentiment intensities were estimated for blue and red states in two-week sliding windows, and a ratio of average sentiment intensities in blue versus red state was estimated.

IV. DISCUSSION

We are aware of some limitations to the present work. We are aware that location-tagged tweets are a small subset of the overall tweets. We are aware that Twitter users are not a fair representation of the general population of the United States [12]. We are aware that the COVID-19 related lexicon is a changing definition on its own, and a variety of topics are politicized and polarized. We are aware of the limitation of sentiment analysis. For example, we expect similar sentiment scores for negative tweets about mask requirement and negative tweets about anti-mask behaviors. We are aware that slang, sarcasm, and other cultural language particularities may be challenging for accurate sentiment intensity estimation. We are aware of the potential interwind of COVID-19 tweets with other social and political events. For example, the Black Life Matters movement intensified during May 2020. We are also aware that the partition of the states into two categories of blue and red ones is over-simplified.

One point of intrigue in this work is the evident negative correlation between Twitter sentiment and the popular vote for either of the major parties in prolonged periods of 2020. There did appear on the heatmap to be a slightly greater correlation between the sentiments and popular vote for the Republican party, which would be understandable given that it was the party in power during this time, suggesting that people who publicly voiced approval for the state of the country at the time would vote in a way to keep it the same.

Looking at both the heatmap and time series charts, there are also two points in which the sentiment shows a much greater correlation with Democrat voters, so it could be valuable to look closer at the data from the beginning of the data to find out what caused such a significant drop off as well as what caused the spike in May. The first drop may be attributable to lockdown fatigue, which would make it interesting to analyze possible correlation between the sentiments in that time period to the adherence to stay at home orders or social distancing.

## V. Conclusion

We observed weak correlations between the average sentiment of COVID-19 related tweets and popular votes in the 2020 election of the United States. We observed that COVID-19 tweet were more positive in blue states than in red state during the beginning of the pandemics. We found that sentiments in the red states gradually become more positive during the summer of 2020 and persisted until the election day. We observed this shifting trend using both a heatmap and a ratio-based comparative analysis. Future work will be required to investigate the possible sources of these changes and their overall implications for the role of social media and the SARS-CoV-2 pandemic in the politics of the United States.

## Acknowledgment

We thank the support of award #1663105 and #1852042 from the National Science Foundation (NSF) and a capacity building project from Department of Defense Cyber Scholarship program (DoD CySP) of the United States.

## References

[1] Gusmano, M.K., et al., *Partisanship in initial state responses to the COVID‐19 pandemic.* World Medical & Health Policy, 2020. 12(4): p. 380-389.

[2] Harvey, N., *Behavioral fatigue: Real phenomenon, naïve construct, or policy contrivance?* Frontiers in Psychology, 2020: p. 2960.

[3] Liu, H., et al., *COVID-19 information overload and generation Z's social media discontinuance intention during the pandemic lockdown.* Technological Forecasting and Social Change, 2021. 166: p. 120600.

[4] Burel, G., T. Farrell, and H. Alani, *Demographics and topics impact on the co-spread of COVID-19 misinformation and fact-checks on Twitter.* Information Processing & Management, 2021. 58(6): p. 102732.

[5] Battaglio, S., *When red meant Democratic and blue was Republican. A brief history of TV elecgoral maps.*, in *Los Angeles Times*. 2016.

[6] Wang, H., et al. *A system for real-time twitter sentiment analysis of 2012 us presidential election cycle*. in *Proceedings of the ACL 2012 system demonstrations*. 2012.

[7] Lamsal, R., *Design and analysis of a large-scale COVID-19 tweets dataset.* Applied Intelligence, 2021. 51(5): p. 2790-2804.

[8] Staff, A., *A Timeline of COVID-19 Developments in 2020.* American Journal of Managed Care, 2021. 1.

[9] Wasserman, D., et al., *2020 Popular Vote Tracker.* 2021: Cook Political.

[10] Hutto, C. and E. Gilbert. *Vader: A parsimonious rule-based model for sentiment analysis of social media text*. in *Proceedings of the international AAAI conference on web and social media*. 2014.

[11] Waskom, M.L., *seaborn: statistical data visualization.* Journal of Open Source Software, 2021. 6(60): p. 3021.

[12] Mislove, A., et al., *Understanding the Demographics of Twitter Users.* Proceedings of the International AAAI Conference on Web and Social Media, 2021. 5(1): p. 554-557.